# Modeling of Mixed Decision Making Process

Nesrine BEN YAHIA, Narjès BELLAMINE and Henda BEN GHEZALA
RIADI-GDL laboratory
National School of Computer Sciences
Manouba, Tunisia
Nesrine.benyahia@ensi.rnu.tn

*Abstract*—Decision making whenever and wherever it is happened is key to organizations success. In order to make correct decision, individuals, teams and organizations need both knowledge management (to manage content) and collaboration (to manage group processes) to make that more effective and efficient. In this paper, we explain the knowledge management and collaboration convergence. Then, we propose a formal description of mixed and multimodal decision making (MDM) process where decision may be made by three possible modes: individual, collective or hybrid. Finally, we explicit the MDM process based on UML-G profile.

## I. INTRODUCTION

Individuals and groups, within organisations, cooperate by producing, manipulating and organizing knowledge, and by building and refining new collective knowledge. Organisations increasingly see their intellectual capital as strategic resources that must be managed effectively to achieve competitive advantage. This capital consists of the knowledge held in the minds of its members, embodied in its procedures and decision making processes, and stored in its repositories.

Subsequently, it should be useful for KM systems and Collaboration systems to integrate both kinds of capabilities into a single collaborative-and-knowledge based system to support joint efforts towards a goal [1].

Decision making is one of the critical processes where we need both knowledge management (that focuses on creation, storage, sharing and use of knowledge) and collaboration (that focuses on cooperation, communication, coordination and coproduction) to make that more effective and efficient.

This paper aims to explicit step-by-step the multimodal decision making (MDM) process at three levels (individual, collective and hybrid) and is organized as follows; we start with a brief overview of the literature on collaborative knowledge management. In section three, we propose formal description of MDM process. Finally, section four presents our model of MDM process basing on the proposed formal description and UML-G profile.

## II. ON THE CONVERGENCE OF COLLABORATION AND KNOWLEDGE MANAGEMENT

To facilitate the understanding of collaborative knowledge management construct, we start our study in this section by defining knowledge management and collaboration. Finally, we examine their convergence.

### A. Knowledge Management

Knowledge is a somewhat elusive concept [2] having many different definitions. For example, [3] describes knowledge under five different perspectives: state of mind, object, process, access to information, and capability. Knowledge is considered as the sum of information in the context that is dependent on the social group creating it [4]. In this paper, we adopt the definition proposed in [5]: Knowledge as *a fluid mix of framed experience, values, contextual information, and expert insight that provides a framework for evaluating and incorporating new experiences and information.*

Knowledge management is largely considered as a process that combines various activities. Following a literature study of KM practices, [1] synthesize generic KM activities as follow: Create (develop new understandings), Collect (acquire and record knowledge), Organize (establish relationships and context so that collected knowledge can be easily accessed), Deliver (share knowledge), Use (bear knowledge on a task).

In this paper we consider special kind of KM which is experience management defined in [6] as *the dissemination of specific knowledge situated in a particular problem-solving context.*

In different viewpoints on KM, many classifications of KM approaches can be distinguished, among them we keep the classification which is originally proposed by [7] and recently adopted by [8]. These authors distinguish two approaches of KM: codification versus personalization.

Codification approaches consider that Knowledge can be articulated, codified and disseminated in the form of documents, drawings and best practices. Knowledge can be shared via knowledge base or repository.

Personalization approaches consider that Knowledge is personal in nature and very difficult to extract from people. Knowledge can be shared via interaction between participants.

## B. Collaboration

We adopt two perspectives of collaboration: the first one is that collaboration may be seen as the combination of communication, coordination and cooperation [9] and [10]. Communication is related to the exchange of messages and information among people, coordination is related to the management of people their activities and resources, and cooperation is related to the production taking place on a shared space. The second one is that collaboration is s a coordinated activity where the attempt is to construct and maintain a shared conception of a problem [11].

Computer Supported Cooperative Work (CSCW) is considered as an attempt to understand the nature and characteristics of collaborative work [12]. It indicates the scientific study and theory of how people work together, how the computer and related technologies affect group behavior, and how technology (groupware) can best be designed and built to facilitate group work [9].

## C. Collaborative Knowledge Management (CKM)

KM and collaboration are complementary [13]. We identify several terms that are used to denote the convergence of KM and collaboration: collaborative knowledge management can be considered as a process of collective resolution of problems where it is useful to memorize the process of making collective decision and to structure the group interactions to facilitate problem solving and sharing of ideas [14].

Collaborative knowledge building can be defined as a sequential social process in which participants' co-construct knowledge through social interactions and that incorporates multiple distinguishable phases that constitute a cycle of personal and social knowledge building [15].

Collaborative knowledge sharing can be supposed as the development of a shared knowledge repository via groupware where conflicts and divergent opinions are an important source to aliment it and their resolution generates new collaborative knowledge [16].

Collaborative knowledge construction can be explained as a learning process where collaborative groups built on the new ideas offered by others, expressing agreement, disagreement, and modifying the ideas being discussed [17].

Finally Collaborative knowledge creation can be considered as the ability to increase the knowledge base or repository, to develop new capabilities and to enhance existing capabilities through combination and knowledge exchange [18].

## D. Proposal of generic framework of CKM

In order to synthesize the convergence of collaboration and KM, we propose a generic framework based on three spaces:

- Collaboration Space: concerns the management of cooperative tasks and it covers the communication, coordination, coproduction and awareness. By awareness, participants may be conscious and may obtain feedback from their actions and from the actions of their companions by means of Meta knowledge. Awareness Meta knowledge consider who (participants), what (collective knowledge), how (management manner), when and where (time and space) of this process.

- KM Space: concerns the management of (collective) knowledge and it covers the strategies of KM: codification (which considers the computer human interaction) and personalization (which considers the computer human-human interaction).

- Actors Space: concerns the management and the representation of the different actors and their roles. There are three types of actors: individuals (that work independently in a private context), groups (dependent individuals that work together in the shared context and engaged to achieve a common goal) and organizations (groups that work collectively and collaboratively to achieve the organisational goals).

Then, the development and the creation of the knowledge base must be considered to enhance the collaborative management [16]-[18].

In addition, we join the idea of [16] on managing two types of knowledge repository: private and shared.

- In a private context and workspace, individual can administer a private knowledge base which is only accessible by him/her and represents the private view of the shared one.

- In the shared context and workspace, there is a unique and public knowledge base which is accessible to everyone.

Since individuals can manage two types of knowledge memory (private and public), they can work in private or shared context by:

- Externalization: when individual store knowledge in his private knowledge memory, this knowledge is converted from tacit to explicit in the private context.

- Publication: when individuals make public some externalized knowledge and store them in the shared knowledge memory so knowledge are moved from private to shared context, or when groups add knowledge to the shared memory in the public context.

- Internalization: when individuals or groups use knowledge from the shared knowledge memory so knowledge is converted from explicit to tacit knowledge however it is moved from shared to private context only for individual internalization.

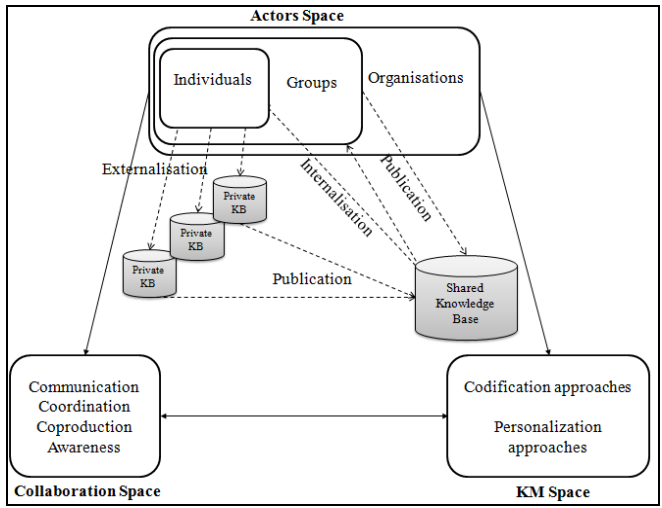

Figure 1. A generic framework of CKM.

### III. MULTIMODAL DECISION MAKING

#### A. MDM specification

Our aim here is to focus on multimodal decision making process. In this paper, specifically, we consider MDM process as a collaborative knowledge management process where knowledge represent experiences, Knowledge management represents creation, organization and dissemination of specific knowledge situated in a particular problem solving context, collaborative knowledge management represents process of problem resolution (based on four phases as it is proposed by [19] and revisited by [20]: intelligence, design, choice and review. In the intelligence phase, the problem is identified. In the design phase, the proposed alternatives or solutions are generated. In the choice phase a solution is selected. Finally, in the revision phase the choice is revised and an intelligent feedback permits to correct errors), knowledge base represents cases base where we store experiences (each case represents a problem, different alternatives to solve the problem and the final decision), codification KM represents the computer human aspect where actors interact with their computer (via their private or public KB) to solve their problem, personalization KM represents the computer human-human aspect where actors interact together with their computers (synchronous or asynchronous, located or distributed) to solve their (collective) problem.

The essential property of MDM here, we argue, is that it enables three modes of problem resolution and decision making: individual mode (computer human interaction), collective mode (computer human-human interaction) and hybrid mode (navigation between the two previous cases).

In this paper, we argue also that one of the interesting points to be considered in the MDM process is the actors' dynamicity as it represents a process by which individuals formulate the problem (together), generate and evaluate solutions (together) and make decision (together).

Thus, we characterize MDM process by actor oriented perspective where actors guide the MDM process and orient the resolution mode in all phases (problem formulation, solutions generation or decision making).

Dynamicity of actors is useful especially when we talk about hybrid mode resolution (when we navigate between the two modes individual and collective). For example, if one individual has a problem he can choose to construct and formalize it alone or with others to help him. After problem conception (individual or collective), the same individual can generate solutions alone or with others. After solutions conception (individual or collective), he can choose one alternative alone or with others.

Accordingly, we distinguish between three types of actors: problem-constructor, solution-constructor and decision-maker:

- Problem-formulator: identify, formulate and structure the problem. His aim is to find out the problem.

- Solution-generator: generate and propose solutions. His aim is to generate alternatives, set criteria and scenarios to evaluate alternative.

- Decision-maker: select and choose one alternative. His aim is to choose alternative(s) and determine the outcome of chosen alternatives.

In the problem conception, MDM process can be characterized as single problem-formulator or multi problem-formulator. In the solutions conception, MDM process can be characterized as single solution-generator or multi solution-generator. In the selection, MDM process can be characterized as single decision-maker or multi decision-maker.

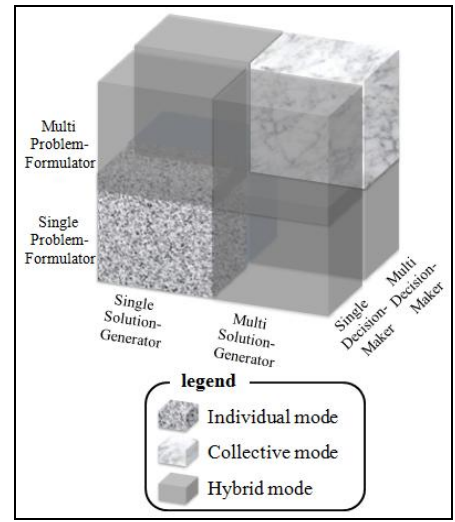

Figure 2. MDM: dynamicity of actors' perspective.

The individual mode of problem resolution corresponds to the case single problem-formulator, single solution-generator and single decision-maker.

The collective mode of problem resolution corresponds to the case multi problem-formulator, multi solution-generator and multi decision-maker.

The navigate or hybrid mode of problem resolution corresponds to the rest cases.

## B. MDM Formal description

In order to formalize and organize the MDM process, we propose a formal description using vector model as follow: $<A,P,S,D>$.

Where A contains actors participating to the different activities of the MDM, and so $A=<A_p, A_s, A_d>$:

$A_p$ represents the actors that participate to the problem conception, $|A_p|$ = nbr(problem-formulators).

$A_s$ represents the actors that participate to the solutions conception, $|A_s|$ = nbr(solution-generators).

$A_d$ represents the actors that participate to the decision making, $|A_d|$ = nbr(decision-makers).

$\forall a \in A$, $a = <a_1, a_2,..., a_n>$ where $a_i$ represents attributes of a so the profile of a.

Then, P represents the problem description: $P = <p_1, p_2,..., p_m>$ where $p_i$ represent attributes of P.

S represents the solutions space: $S = S_c \cup S_p$, and $\forall s \in S$, $s = <s_1, s_2,..., s_m>$ where $s_i$ represent attributes of s.

$S_c$ represents the different solutions or alternatives that are generating from the private and/or knowledge bases (via codification approach of knowledge management) and $S_p$ represents the different solutions or alternatives that are generating and proposed by solution makers (via personalization approach of knowledge management).

If $|A_s|$ = 1 (only the problem maker will solve his own problem using computer human interaction without interacting with others) then $S_p = \emptyset$ and $S = S_c$.

But, if $|A_s|$ > 1 (group of users will solve shared problem using computer human-human interaction) then $S_p \neq \emptyset$.

Finally, D represents the final decision ($D \in S$), it corresponds to the output of MDM process which is individual if $|A_d|$ = 1 and collective if $|A_d|$ > 1.

## C. MDM process model

In order to explicit and explore step-by-step the flow control of MDM process, we combine the use of UML 2.0 activity diagram and UML-G profile.

In activity diagram UML 2.0, swimlanes are manipulated to group activities performed by the same actor. So, in our case we need three swimlanes (the first one contains activities of problem-formulator, the second one contains activities of solution-generator and the third one contains activities of decision-maker). Guards of decision are based on the proposed formal description of MDM.

In order to model shared element, we use UML-G introduced in [21] as an UML profile for modeling groupware. The stereotype <<shared>> is introduced and can be applied to any UML model element. Instances of elements marked as <<shared>> are potentially accessible from all users. <<sharedRole>> and <<sharedActor>> are also introduced as separate stereotypes in order to mark their special meaning. With <<sharedRole>> roles in cooperative sessions can be shared between several actors and with <<sharedActor>> actors in cooperative sessions can take several roles. Similarly, they introduce the stereotype <<sharedActivity>> to denote all collective activities.

In MDM process now, $\forall a \in A$, a is marked with both stereotypes <<sharedRole>> and <<sharedActor>> as a may respectively share his role with others (i.e. if $a \in A_p$ and $|A_p|$> 1 and/or if $a \in A_s$ and $|A_s|$> 1 and/or if $a \in A_d$ and $|A_d|$> 1) and have many roles at the same time (i.e. $a \in A_p \cap A_s$ or $a \in A_p \cap A_d$ or $a \in A_s \cap A_d$ or $a \in A_p \cap A_s \cap A_d$).

As shown in figure 3, MDM process contains the following activities:

Formulate P: represents the problem conception and formulation which is individual if $|A_p|$= 1(the problem-formulator formalizes his own problem alone) and collective if $|A_p|$> 1(the problem-formulators share the problem conception and make consensus of problem representation).

Generate S: represents the solutions conception which is individual if $|A_s|$ = 1and collective if $|A_s|$> 1. In the first case, solution-generator uses his own private KB and public KB to construct $S_c$ by finding out similar stored problem and consult their resolution. In the second case, there are two possible ways: the solution-generators co-construct $S_P$ by proposing directly their alternatives without using the KB or they co-construct $S_P$ and $S_c$ using private and public KB.

Evaluate S: represents the selection of criteria and strategy to evaluate proposed alternatives (by voting, by ordering or by affecting priorities).

Make D: represents the decision making which is individual if $|A_d|$= 1and collective if $|A_d|$> 1. Individual decision will be stored in individual knowledge base and collective decision (marked with <<Shared>> stereotype) will be stored in shared knowledge base.

Maintain: represents maintenance and review of final (collective) decision.

## IV. CONCLUSION

In this paper, we introduce Multimodal Decision Making process as collaborative knowledge management process that covers three modes of problem resolution and decision making (individual, collective and hybrid) and is characterized by dynamicity of actors where we separate between three types of actor (problem-formulator, solution-generator and decision-maker). In order to explicit step-by-step the MDM process, we combine the use of a proposed formal model with UML-G to elaborate an activity diagram of this process.

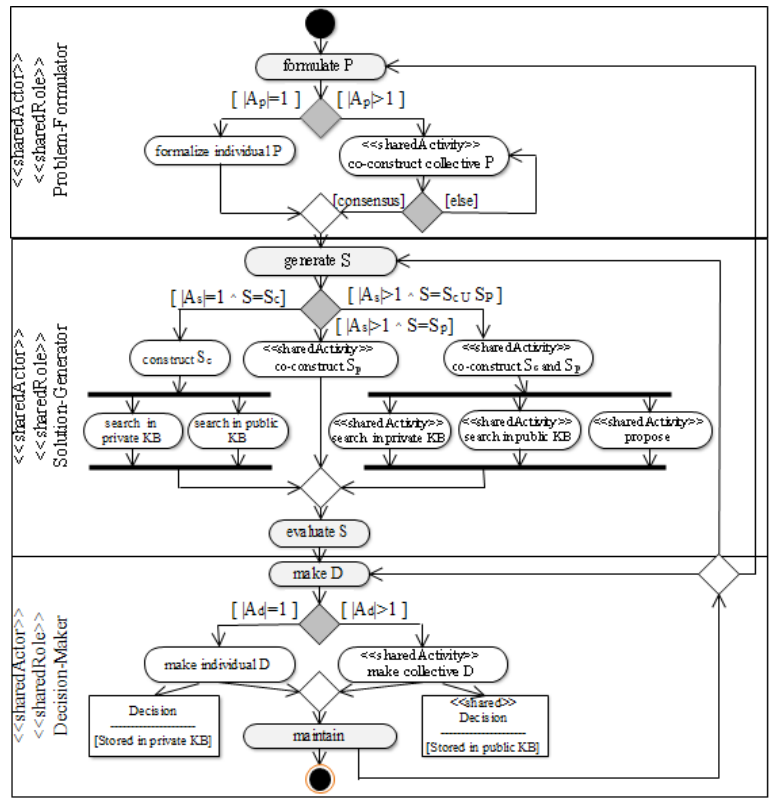

Figure 3. MDM process: UML activity diagram.